\documentclass[sigconf,noacm]{acmart}
\providecommand{\event}{ICSE JAW 2026}

\usepackage{amsmath, amssymb, amsthm, amsfonts}
\usepackage{graphicx}
\usepackage{threeparttable}
\usepackage{tabularx,array,booktabs}
\usepackage{xurl}
\usepackage{url}
\usepackage{threeparttable}
\usepackage{tablefootnote}
\usepackage{xcolor}
\newcolumntype{Y}{>{\raggedright\arraybackslash}X}

\newtheorem{theorem}{Theorem}[section]

\newtheorem{definition}[theorem]{Definition}
\theoremstyle{definition}

\newcommand{\powerset}{\mathcal{P}}

\newcommand{\ep}{\epsilon}
\newcommand{\blank}{\sqcup}

\newcommand{\Tr}{\mathsf{Tr}}
\newcommand{\Runs}{\mathsf{Runs}}
\newcommand{\supp}{\mathsf{supp}}
\newcommand{\Abs}{\alpha}

\usepackage{iftex}
\usepackage{hyperref}
\usepackage{tikz}
\usetikzlibrary{automata, positioning, arrows}
\usetikzlibrary{shapes.geometric}
\usepackage{booktabs}

\newcommand{\newtext}[1]{\textcolor{black}{#1}}

\usepackage{iftex}
\ifpdftex
  \usepackage[T1]{fontenc}
  \usepackage{underscore}
\else
  \usepackage{breakurl}
\fi

\setcopyright{none}
\copyrightyear{}
\acmYear{}
\acmDOI{}
\acmISBN{}
\acmConference{}
\acmBooktitle{}
\acmPrice{}

\title{Are Agents Probabilistic Automata? A Trace-Based, Memory-Constrained Theory of Agentic AI}

\author{
\href{mailto:roham.koohestani@jetbrains.com}{Roham Koohestani}$^{\dagger\ddagger}$,
\href{mailto:ziyou.li@tudelft.nl}{Ziyou Li}$^{\ddagger}$
\href{mailto:anton.podkopaev@jetbrains.com}{Anton Podkopaev}$^{\dagger\S}$, 
\href{mailto:m.izadi@tudelft.nl}{Maliheh Izadi}$^{\ddagger}$
}

\affiliation{%
  \institution{\small
    $^\dagger$JetBrains Research, Amsterdam, the Netherlands \quad
    $^\ddagger$Delft University of Technology, Delft, the Netherlands \\
    $^\S$Constructor University, Bremen, Germany \quad
  }
  \country{}
}

\def\titlerunning{Are Agents Probabilistic Automata?}
\def\authorrunning{Koohestani et al.}
\begin{document}

\begin{abstract}
This paper studies standard controller architectures for agentic AI and
derives automata-theoretic models of their interaction behavior via
trace semantics and abstraction.
We model an agent implementation as a finite control program augmented
with explicit memory primitives (bounded buffers, a call stack, or
read/write external memory) and a stochastic policy component
(e.g., an LLM) that selects among architecturally permitted actions.
Instead of equating the concrete agent with a deterministic acceptor,
we treat the agent-environment closed loop as inducing a probability
distribution over finite interaction traces.
Given an abstraction function $\Abs$ from concrete configurations to a
finite abstract state space, we obtain a probabilistic trace language
and an abstract probabilistic transition model $M_{\Abs}$ suitable for
probabilistic model checking.

Imposing explicit, framework-auditable restrictions on memory access and control
flow, we prove that the \emph{support} of the resulting trace language is
regular for bounded-memory controllers, context-free for strict
call-return controllers, and recursively enumerable for controllers
equipped with unbounded read/write memory.
These correspondences allow the reuse of existing verification
methods for finite-state and pushdown systems, and they delineate precisely
when undecidability barriers arise.
The probabilistic semantics leads to quantitative analyses such as:
what is the probability of entering an unsafe abstract region, and how can
we bound this probability in the presence of environment nondeterminism.
\end{abstract}

\maketitle

\section{Introduction}
\label{sec:introduction}
The rapid evolution of artificial intelligence has transitioned from reactive single-turn systems to proactive autonomous entities known as agentic AI. 
These systems are designed to perceive their environment, formulate plans, and execute multi-step tasks using tools to achieve goals with minimal human intervention~\cite{he_llm-based_2025}. 
Although often described in terms of cognitive capabilities like \textit{reasoning} and \textit{planning}, 
many agent frameworks implement a control loop that can be described as a labeled transition system once the environment and tool interfaces are made explicit~\cite{wu_stateflow_2024}.

\newtext{
This paper presents a framework that classifies agents by the persistent memory interfaces they enforce and by the resulting interaction traces. The classification has a qualitative layer and a quantitative layer. Qualitatively, under checkable architectural constraints, the \emph{support} of the induced trace language is regular (bounded persistent memory), context-free (stack-scoped call-return), or recursively enumerable (unbounded read/write memory). Quantitatively, the closed loop induces a probability distribution over traces; after abstraction, this yields a finite probabilistic transition model (a Markov chain or MDP) for bounded-memory controllers and pushdown/recursive probabilistic models for stack-scoped controllers.
}

\vspace{-1em}

\textbf{Goal and Contribution: }
This work bridges agent engineering practice and automata-theoretic reasoning by focusing on architectural constraints that are auditable at the framework level. Our contributions are:
(1) \textbf{A \newtext{trace-}centric classification:} we define agent classes in terms of enforced persistent memory operations (finite-state, stack-scoped, or unbounded read/write memory) \newtext{and based on the produced traces of the agents}.
(2) \textbf{A right-sizing guideline:} we provide a decision procedure at the architecture level for selecting the weakest controller class consistent with task requirements.
(3) \textbf{Research Roadmap:} We outline a series of research paths to explore which focus on formal verification and testing, as well as practical engineering issues related to right-sizing.

In the following section, we provide the necessary background (\autoref{sec:bg}) and define our framework (\autoref{sec:approach}). Sections \ref{sec:reg-prob}--\ref{sec:tca} establish the correspondence between architectural memory constraints and their respective automata classes. We extend these results to multi-agent systems in \autoref{sec:mas}. Finally, we operationalize this theory into a right-sizing procedure (\autoref{sec:rightsizing}) and a case study (\autoref{sec:casestudy}), concluding with a discussion in \autoref{sec:discussion}.

\section{Background and Related Work}
\label{sec:bg}

\textbf{Agentic AI Architectures.}
Modern agentic AI systems are composed of several key components that work in a continuous operational loop. This loop, often a variant of the Sense-Plan-Act cycle,\footnote{E.g., \url{https://motion.cs.illinois.edu/RoboticSystems/AnatomyOfARobot.html}} dictates how the agent interacts with its environment.
\textbf{Perception:} The agent takes in information from its environment, which could be user input, sensor data, or the state of a software system.
\textbf{Reasoning (LLM):} Large Language Models (LLM) are often used as the \textit{brain} of the agent. These are tasked with processing the perceived information, maintaining a model of the world, and deciding on the next course of action.
\textbf{Action Layer:} The agent executes actions through a set of available tools or APIs, which can modify the environment or its internal state.
\textbf{Memory:} This is the component that stores information about past states, actions, and observations. The structure and capacity of this memory are central to our thesis.

Operational loop variants like ReAct (Reason+Act)~\cite{yao_react_2023}, StateFlow~\cite{wu_stateflow_2024}, and SMOT (Step-wise Memory-augmented Task-oriented)\newline ~\cite{liu_smot_2023} all represent different strategies for managing this cycle, but fundamentally operate as state-transition systems.

Our classification targets the controller and its persistent memory interface. We assume tools are invoked through a finite action alphabet $\Gamma$ and that tool outputs are returned as perceptions via $\Sigma$ after tokenization. While practical tool interfaces often accept arguments from infinite domains (e.g., arbitrary text strings or real numbers), we assume for this structural classification that such arguments are either discretized via an abstraction layer (e.g., treating all invalid inputs as a single symbol) or generated by the model (and are therefore finite. Unless stated otherwise, we treat tools and the environment as nondeterministic but computable processes; the classification does not claim to bound the computational power of external oracles, only the controller’s ability to store and reuse information across steps.

\textbf{Trace Semantics and Trace Languages.}
A standard way to model interactive systems is as labeled transition systems, where executions induce \emph{traces}, that is, finite or infinite sequences of observable events~\cite{baier_katoen_modelchecking_2008}. Trace semantics is used as a linear-time abstraction of behavior and supports property classes such as safety (prefix-closed) and reachability. In controller-style agent architectures, an interaction trace can be taken as a sequence of (perception, action) pairs, or as a sequence of pre/post tool-invocation snapshots paired with tool calls, depending on the chosen observation interface.
This paper uses traces as the semantic object for two reasons. First, traces provide a uniform interface across heterogeneous agent frameworks whose internal control
graphs and tool APIs differ. Second, traces align with verification workflows that interpret a controller plus environment as generating a set (or distribution) of observable executions, and then check reachability-style properties over those executions.

\textbf{Probabilistic Transition Models for Stochastic Controllers.}
When a controller’s policy is stochastic (for example, due to sampling in an LLM) and tools exhibit randomness or uncertainty, the closed-loop system induces a probability distribution over traces. A common modeling boundary separates (i) \emph{purely probabilistic} evolution, represented by Markov chains, and (ii) \emph{mixed nondeterministic and probabilistic} evolution, represented by Markov decision
processes (MDPs). MDPs model nondeterministic choices resolved by a scheduler (or adversary) and probabilistic choices resolved by transition probabilities
~\cite{puterman_mdp_1994,baier_katoen_modelchecking_2008}.
Another closely related formalism is the \emph{probabilistic automaton} model that combines nondeterminism with probabilistic transitions. In verification, this is often treated as an MDP-style transition system in which each state enables a set of probability distributions over successors~\cite{segala_lynch_probabilistic_simulations_1995}.

\newtext{
These models match agent settings where tool outcomes are not fully specified or where abstraction introduces uncertainty, while still supporting quantitative questions such as maximum probability of reaching an unsafe region.
}

\textbf{Probabilistic Model Checking.}
Probabilistic model checking analyzes quantitative properties of Markov chains and MDPs, including probabilities of reachability, expected accumulated costs, and temporal-logic specifications such as PCTL and related logics. Algorithms for reachability in Markov chains reduce to solving linear equations; for MDPs they reduce to fixed-point characterizations that compute extrema over all schedulers~\cite{baier_katoen_modelchecking_2008}. Modern probabilistic model checkers implement these analyses and expose multiple interpretations of nondeterminism (for example, worst-case and best-case reachability). This paper’s abstraction-based modeling step is intended to produce precisely the kind of finite probabilistic model that these tools consume~\cite{dehnert_storm_2017}.

\textbf{Recursive and Pushdown Models.}
Hierarchical controllers with call-return structure are naturally modeled by pushdown systems. When probabilistic choices are present, one obtains probabilistic pushdown models, which can also be presented as recursive stochastic transition systems. A representative formalization is via recursive Markov chains, where each procedure is a component with entry and exit nodes and control can invoke components recursively~\cite{etessami_yannakakis_rmc_2005}. These models provide the standard bridge between (i) context-free control structure and (ii) quantitative reachability analysis. They are therefore the appropriate background for modeling the quantitative behavior of stack-scoped agent controllers.

\textbf{Abstraction and Refinement.}
Direct verification of realistic controllers is limited by state explosion and by the need to model tools and environments at an appropriate level of detail. Abstraction addresses this by mapping concrete configurations to a finite abstract state space, and then verifying properties on the induced abstract model. For nondeterministic or probabilistic systems, abstraction must account for both sources of uncertainty, and coarse abstractions can introduce spurious behaviors that do not correspond to any concrete execution. Counterexample-guided abstraction refinement (CEGAR) is a standard methodology that iteratively refines abstractions using counterexamples produced by verification~\cite{clarke_cegar_2000}.
In probabilistic settings, counterexamples may need to represent sets of traces or scheduler choices, and refinement may separate abstract states to eliminate spurious high-probability paths~\cite{dehnert_abstraction_probabilistic_2017}.

\textbf{Formal Methods in AI Systems.}
The application of formal methods to AI is not new. Classical multi-agent systems have been analyzed using model-checking tools like AJPF~\cite{dennis_mcapl_2018} and MCMAS~\cite{sail_mcmas_2006}. In the context of LLMs, techniques such as grammar-constrained decoding~\cite{park_grammar-aligned_2024}, Linear Temporal Logic (LTL) monitors~\cite{cohen_end--end_2024}, and frameworks like Formal-LLM~\cite{li_formal-llm_2024} have been used to enforce certain properties on the output. However, a significant gap remains: there is no unifying, memory-based classification that links the architectural patterns of modern agentic AI to the fundamental decidability and complexity results from automata theory. This paper aims to fill that gap.

\section{A Formal Theory for Agentic AI}
\label{sec:approach}
To develop a formal theory for agents, we must first define the standard automata for establishing equivalence. We subsequently present our claims for equivalence.

\subsection{Standard Automata Definitions}
\label{app:defs}
The definitions that follow are standard in the theory of computation and are consistent with the established literature. 

\paragraph{Finite Automaton (FA)} is a 5-tuple $M=(Q, \Sigma_{in}, \delta, q_0, F)$ where:
\begin{itemize}
    \item $Q$ is a finite set of states.
    \item $\Sigma_{in}$ is a finite set of input symbols, called the alphabet. 
    \item $\delta: Q \times \Sigma_{in} \rightarrow Q$ is the transition function.
    \item $q_0 \in Q$ is the start state.
    \item $F \subseteq Q$ is the set of accept states.
\end{itemize}

\textbf{A note on perceptions and finiteness.}
We model a controller step as consuming a \emph{single} perception symbol $a \in \Sigma_{in}$ produced by a tokenizer $\tau$ applied to raw observations. This abstracts away variable-length raw observations by treating tokenization and observation packaging as part of the environment. The controller is \emph{finite-state} only when (i) the controller state set is finite and (ii) any persistent memory accessible across steps is constant-bounded and encoded into the finite control. If the architecture exposes a persistent store whose size can grow with execution, the controller is not finite-state regardless of tokenization.

It is well-established that for every NFA, an equivalent DFA can be constructed using the subset construction method; thus, DFAs and NFAs recognize the same class of languages, being the regular languages.
We define the extended transition where for $M=(Q,\Sigma_{in},\delta,q_0,F)$, define $\hat\delta:Q\times\Sigma_{in}^\ast\to Q$ is given by
$\hat\delta(q,\epsilon)=q$ and $\hat\delta(q,xa)=\delta(\hat\delta(q,x),a)$ for $x\in\Sigma_{in}^\ast$, $a\in\Sigma_{in}$.

\paragraph{Pushdown Automaton (PDA)} is a 6-tuple $P=(Q, \Sigma_{in}, \Gamma, \delta, q_0, F)$:
\begin{itemize}
    \item $Q,\Sigma_{in},q_0 \in Q,F \subseteq Q$ defined similarly to FAs.
    \item $\Gamma$ is a finite stack alphabet.
    \item $\delta: Q \times \Sigma_{in,\ep} \times \Gamma_{\ep} \rightarrow \powerset(Q \times \Gamma_{\ep}^*)$ is the transition function, where $\Sigma_{in,\ep} = \Sigma_{in} \cup \{\ep\}$ and $\Gamma_{\ep} = \Gamma \cup \{\ep\}$.
\end{itemize}
PDAs recognize the class of context-free languages ~\cite{sipser_introduction_2012}. In terms of acceptance and configurations,
we use \emph{acceptance by final state}. A configuration is $(q,w,\alpha)\in Q\times\Sigma^\ast\times\Gamma^\ast$.
We write $(q,aw,\gamma\alpha)\vdash(q',w,\alpha')$ if $(q',\alpha')\in\delta(q,a,\gamma)$ (with $\epsilon$-moves when $a=\epsilon$ or $\gamma=\epsilon$).

\paragraph{Turing Machine} A standard, single-tape TM is a 7-tuple $M=(Q, \Sigma_{in}, \Gamma, \delta, q_0, q_{\text{accept}}, q_{\text{reject}})$ where:
\begin{itemize}
    \item $Q,\Sigma_{in},q_0 \in Q$ defined similarly to FAs.
    \item $\Gamma$ is the tape alphabet, where $\Sigma_{in} \subseteq \Gamma$ and $\blank \in \Gamma$.
    \item $\delta: Q \times \Gamma \rightarrow Q \times \Gamma \times \{L, R\}$ is the transition function.
    \item $q_{\text{accept}} \in Q$ is the accept state.
    \item $q_{\text{reject}} \in Q$ is the reject state, where $q_{\text{accept}} \neq q_{\text{reject}}$.
\end{itemize}
TMs recognize the class of recursively enumerable languages.

To prove correspondence, the agent architectures described in the paper must be abstracted into a formal model. Notably, we use acceptor models only to characterize the \emph{support language} of feasible traces under architectural constraints. Probabilities enter through the induced trace distribution and the abstract model $M_{\Abs}$, not through probabilistic acceptance semantics of automata.

\subsection{Agentic Trace Definitions}
\begin{definition}[Concrete runs and trace projection]
\label{def:runs-traces}
Let a controller $A$ interact with an environment in discrete steps.
Let $C$ be the set of concrete configurations (state, memory contents, and any fixed bookkeeping).
A \emph{run} is a (finite) sequence
$
\rho = c_0 a_0 u_0 c_1 a_1 u_1 \cdots c_n
$
where each $c_i \in C$, each $a_i \in \Sigma$ is the packaged perception produced by the environment, and each $u_i \in \Gamma$ is an action.
We write $\pi(\rho) = (a_0,u_0)(a_1,u_1)\cdots(a_{n-1},u_{n-1}) \in (\Sigma\times\Gamma)^*$.
\end{definition}

\newtext{
\begin{definition}[Probabilistic trace semantics]
\label{def:prob-trace}
Assume the closed loop of controller, environment, and policy induces a
probability measure $\Pr_A$ over finite runs (equivalently, over finite trace prefixes).
The \emph{probabilistic trace semantics} of $A$ is the pushforward
distribution over traces:
$
\Pr_A^{\Tr}(t)
=
\Pr_A\!\left(\{\rho \in \Runs(A) \mid \pi(\rho)=t\}\right),
t \in (\Sigma \times \Gamma)^{*}.
$
The pair $(\Tr(A), \Pr_A^{\Tr})$ is a probabilistic trace language.
Its support $\supp(\Pr_A^{\Tr})$ is the set of traces with nonzero
probability.
\end{definition}
}

\begin{definition}[State abstraction and abstract traces]
\label{def:alpha}
Let $\Abs : C \to \widehat{S}$ be an abstraction map into a finite set
$\widehat{S}$ of abstract states.
While a concrete run $\rho = c_0 a_0 u_0 c_1 \cdots$ produces a trace $\pi(\rho) \in (\Sigma \times \Gamma)^*$, the abstraction allows us to view this run as a path through abstract states : $\widehat{\rho} = \Abs(c_0) \xrightarrow{a_0/u_0} \Abs(c_1) \dots$.
We can say that the abstraction induces a probabilistic transition model over $\widehat{S}$ (which we define formally in~\autoref{def:abstract-model}).
The \emph{abstract trace semantics} is the distribution over $(\Sigma\times\Gamma)^*$ induced by this abstract model. Note that unlike the state sequence, the trace retains the perceptions $a_i$, as these form the input alphabet of the corresponding automaton. 
\end{definition}

\vspace{-1em}

\newtext{
\begin{definition}[Abstract probabilistic model induced by $\Abs$]
\label{def:abstract-model}
An abstract probabilistic model for $(A,\Abs)$ is a finite transition
model $M_{\Abs}$ over abstract states $\widehat{S}$ that induces a
distribution over abstract traces consistent with $\Pr_{A,\Abs}^{\Tr}$.
When the environment is fully probabilistic, $M_{\Abs}$ can be taken as
a labeled Markov chain.
When the environment includes nondeterministic choices (for example,
unmodeled tool variability), $M_{\Abs}$ is taken as an MDP, or more
generally as a probabilistic automaton in the sense of
nondeterministic choice over probability distributions.
Probabilistic model checking queries are evaluated on $M_{\Abs}$ under
the chosen interpretation of nondeterminism (e.g., worst case, best
case, or strategy-restricted).
\end{definition}
}

\newtext{
In the sections that follow, we establish correspondence by relating each controller class to a standard machine model at the level of \emph{support languages} of traces (via trace-preserving encodings) and by showing how each class induces a suitable abstract probabilistic model $M_{\Abs}$ for quantitative verification.
}

\section{Regular Agents induce Finite-State Probabilistic Models}
\label{sec:reg-prob}

A \textbf{Regular Agent} is a controller architecture whose persistent
memory is constant-bounded and encoded into a finite set of control
states.
The policy component (e.g., an LLM) may be stochastic, and the
environment (including tools) may be probabilistic or may contain
nondeterminism.
As a result, the closed loop induces a probability distribution over
finite interaction traces (Definition~\ref{def:prob-trace}).

This section establishes two correspondences:
(i) the \emph{support} of the induced trace distribution is regular, and
(ii) the \emph{quantitative behavior} is representable by a finite
Markov chain or MDP (depending on how tool/environment uncertainty is
modeled).

\subsection{Architecture constraint}
\label{sec:reg-arch}

\begin{definition}[Regular Agent (architecture)]
\label{def:reg-arch}
A Regular Agent architecture consists of:
(i) a finite set $S$ of control states,
(ii) a finite perception alphabet $\Sigma$ and a finite action alphabet
$\Gamma$, and
(iii) an enforcement mechanism which ensures that all persistent information
that can influence future control decisions is encoded in $S$.
At each step, given a perception $a\in\Sigma$ and state $s\in S$, the
architecture enables a finite set of transitions of the form
$(s,a)\mapsto(s',u)$, where $s'\in S$ and $u\in\Gamma$.
A policy selects among enabled transitions stochastically.
\end{definition}

\paragraph{Exactness of abstraction.}
The correspondence statements are phrased using a finite abstraction
$\Abs:C\to\widehat S$ (Definition~\ref{def:alpha}), where $C$ are
concrete configurations.
For Regular Agents, we require $\Abs$ to preserve the control-relevant
distinctions.

\begin{definition}[Exact abstraction for Regular control]
\label{def:reg-exact}
Let $A$ be a Regular Agent architecture.
An abstraction $\Abs:C\to\widehat S$ is \emph{exact for control} if for
all concrete configurations $c,c'\in C$ with $\Abs(c)=\Abs(c')$ and for
all perceptions $a\in\Sigma$, the sets of architecturally enabled
abstract transitions coincide:
$
\{(\Abs(c_1),u)\mid c \xrightarrow{a/u} c_1\}
=
\{(\Abs(c_1'),u)\mid c' \xrightarrow{a/u} c_1'\},
$
where $c \xrightarrow{a/u} c_1$ denotes a one-step interaction that
consumes perception $a$ and emits action $u$ (with the environment/tool
effects included in the successor configuration).
\end{definition}

\subsection{Regular support via an explicit NFA}
\label{sec:reg-support}

We first build an automaton that generates exactly the feasible
\emph{interaction traces} of the abstracted system.

\begin{definition}[Support NFA for Regular Agents]
\label{def:reg-support-nfa}
Fix a Regular Agent $A$ and an abstraction $\Abs$ with finite codomain
$\widehat S$.
Define an NFA
$
G^{\mathrm{reg}}_{A,\Abs}
=
(\widehat S,\widehat s_0,\Sigma\times\Gamma,\rightarrow,\widehat S),
$
where all states are accepting, $\widehat s_0=\Abs(c_0)$ for the initial
configuration $c_0$, and the transition relation
$\widehat s \xrightarrow{(a,u)} \widehat s'$
holds iff there exists a concrete step from some configuration
$c$ with $\Abs(c)=\widehat s$ that consumes perception $a$, emits action
$u$, and reaches some $c'$ with $\Abs(c')=\widehat s'$.
\end{definition}

The NFA reads symbols in $\Sigma\times\Gamma$ and accepts every finite
trace that is feasible under the architecture (and under the chosen
interpretation of environment feasibility as existential).

\begin{theorem}[Regular support]
\label{thm:reg-support}
Let $A$ be a Regular Agent and let $\Abs$ be an abstraction with finite
codomain $\widehat S$ that is exact for control
(Definition~\ref{def:reg-exact}).
Then the trace support language
$\supp(\Pr^{\Tr}_{A,\Abs}) \subseteq (\Sigma\times\Gamma)^*$
is regular.
In particular,
$
\supp(\Pr^{\Tr}_{A,\Abs}) = L(G^{\mathrm{reg}}_{A,\Abs})
$
for the NFA in Definition~\ref{def:reg-support-nfa}.
\end{theorem}

\begin{proof}[Proof sketch]
Because $\widehat S$ is finite and $\Sigma\times\Gamma$ is finite,
$G^{\mathrm{reg}}_{A,\Abs}$ is a finite automaton.
Exactness ensures that abstract states characterize precisely the
enabled abstract moves, so every feasible abstract trace corresponds to
a path in the NFA and conversely every NFA path corresponds to some
feasible abstract execution.
Regularity follows because the accepted language of a finite automaton
is regular.
\end{proof}
\vspace{-1em}

It must be noted that we assume that the policy assigns nonzero probability to each architecturally enabled transition. Equivalently we could restrict the transition relation in the support generator to those with nonzero probability under the closed-loop semantics.

\subsection{Finite probabilistic model for quantitative queries}
\label{sec:reg-quant}

We now describe the induced quantitative model used for probabilistic
model checking. This depends on how tool/environment uncertainty is
represented.

\begin{definition}[Finite-state probabilistic closed-loop model]
\label{def:reg-mabs}
Fix $A$ and $\Abs$ as above and fix an \emph{environment interpretation}:
either (i) fully probabilistic outcomes or (ii) outcomes containing
nondeterminism.
The induced abstract closed-loop model is
$
M_{\Abs} = (\widehat S,\widehat s_0,\Sigma,\Gamma,\mathcal{T}),
$
where $\mathcal{T}(\widehat s,a)$ is:
\begin{itemize}
  \item a single probability distribution over $(\widehat s',u)\in
  \widehat S\times\Gamma$ in the fully probabilistic case, yielding a
  labeled Markov chain; or
  \item a finite set of distributions over $\widehat S\times\Gamma$ in
  the presence of nondeterminism, yielding an MDP.
\end{itemize}
\end{definition}

\begin{theorem}[Finite probabilistic representation]
\label{thm:reg-quant}
Under a fixed environment interpretation, the model $M_{\Abs}$ in
Definition~\ref{def:reg-mabs} induces the same abstract trace
distribution as $\Pr^{\Tr}_{A,\Abs}$ (Definition~\ref{def:alpha} and
Definition~\ref{def:prob-trace}), under the standard semantics of Markov
chains (probabilistic) or MDPs (nondeterminism resolved by a scheduler).
\end{theorem}

\begin{proof}[Proof sketch]
In each abstract state $\widehat s$ and perception $a$, the architecture
enables a finite set of $(\widehat s',u)$ successors. The policy and
environment interpretation determine either a single probability
distribution or a set of them over these successors.
Because $\widehat S$ is finite, this yields a finite MC/MDP whose stepwise
trace generation matches the pushforward semantics of
Definition~\ref{def:prob-trace} on abstract traces.
\end{proof}

\paragraph{Quantitative verification queries.}
For an unsafe set $\widehat B\subseteq\widehat S$, one evaluates
reachability probabilities such as
$
\Pr_{M_{\Abs}}(\lozenge \widehat B)
$
or\newline
$
\sup_{\sigma}\Pr_{M_{\Abs}}^{\sigma}(\lozenge \widehat B),
$
depending on whether nondeterminism is absent or interpreted via
schedulers $\sigma$.

\section{Context-Free Agents induce Probabilistic Pushdown Models}
\label{sec:cfa-prob}

A \textbf{Context-Free Agent} augments finite control with an
unbounded stack and enforces a stack-scoped persistence discipline
sufficient to model hierarchical control.
As before, we separate:
(i) a qualitative correspondence stating that the \emph{support} trace
language is context-free, and
(ii) a quantitative correspondence stating that the abstract closed loop
is representable by a probabilistic pushdown model (or a pushdown MDP).

\subsection{Architecture constraint}
\label{sec:cfa-arch}

\begin{definition}[Context-Free Agent (architecture)]
\label{def:cfa-arch}
A Context-Free Agent architecture consists of:
(i) a finite set $S$ of control states,
(ii) a finite stack alphabet $Z$,
(iii) finite alphabets $\Sigma$ (perceptions) and $\Gamma$ (actions), and
(iv) an enforcement mechanism restricting persistent memory to a single
stack with LIFO access across steps.
At each step, given control state $s\in S$, perception $a\in\Sigma$, and
top-of-stack symbol $z\in Z_\ep$, the architecture enables transitions
that emit an action $u\in\Gamma$, update the control state to $s'\in S$,
and update the stack by replacing $z$ with a word $\beta\in Z^*$ (push,
pop, or no-op).
Formally, enabled moves are of the shape:
$
(s,z,a)\mapsto(s',\beta,u).
$
A policy selects among enabled moves, possibly stochastically.
\end{definition}

If a strict call-return discipline is desired, one may further partition
actions $\Gamma = \Gamma_{\mathsf{call}}\uplus\Gamma_{\mathsf{ret}}
\uplus\Gamma_{\mathsf{int}}$ and enforce that calls push a designated
return symbol and returns pop it. The correspondence below holds
already for general stack updates, and therefore also for call-return as
a special case.

\subsection{Context-free support via an explicit PDA}
\label{sec:cfa-support}

\begin{definition}[Support PDA for Context-Free Agents]
\label{def:cfa-support-pda}
Fix a Context-Free Agent $A$ and an abstraction $\Abs:C\to\widehat S$
with finite $\widehat S$ that preserves the finite control and any
finite stack information required to determine enabled transitions.
Define a PDA
$
G^{\mathrm{cfa}}_{A,\Abs}
=
(\widehat S,\ \Sigma\times\Gamma,\ Z,\ \delta_G,\ \widehat s_0,\ \widehat S),
$
where $\widehat s_0=\Abs(c_0)$ and $\delta_G$ simulates one abstract
controller step while reading one symbol $(a,u)\in\Sigma\times\Gamma$.
Concretely, $\delta_G(\widehat s,(a,u),z)$ contains $(\widehat s',\beta)$
whenever the architecture permits, from some concrete configuration
$c$ with $\Abs(c)=\widehat s$ and stack-top $z$, consuming perception
$a$, emitting action $u$, moving to some $c'$ with $\Abs(c')=\widehat s'$
and replacing $z$ by $\beta$.
\end{definition}

\begin{theorem}[Context-free support]
\label{thm:cfa-support}
Let $A$ be a Context-Free Agent and let $\Abs$ be a finite abstraction as
in Definition~\ref{def:cfa-support-pda}.
Then
$\supp(\Pr^{\Tr}_{A,\Abs}) \subseteq (\Sigma\times\Gamma)^*$
is context-free, and in particular
$
\supp(\Pr^{\Tr}_{A,\Abs}) = L(G^{\mathrm{cfa}}_{A,\Abs}).
$
\end{theorem}
\vspace{-1em}
\begin{proof}[Proof sketch]
The PDA $G^{\mathrm{cfa}}_{A,\Abs}$ reads interaction symbols
$(a,u)$ and simulates the stack update discipline of the architecture.
Thus it accepts exactly the feasible abstract interaction traces.
Because the accepted language of a PDA is context-free, the support
language is context-free.
\end{proof}

\subsection{Probabilistic pushdown model for quantitative queries}
\label{sec:cfa-quant}

\begin{definition}[Probabilistic pushdown closed-loop model]
\label{def:cfa-mabs}
Fix $A$, $\Abs$, and an environment interpretation as in the Regular
case.
The induced abstract model $M_{\Abs}$ is a probabilistic pushdown model:
a pushdown transition system over control states $\widehat S$ and stack
alphabet $Z$ whose step relation, conditioned on a perception
$a\in\Sigma$, either
(i) assigns a probability distribution over successor triples
$(\widehat s',\beta,u)$ or
(ii) enables a set of such distributions in the presence of
nondeterminism.
In case (ii), the model is a \emph{pushdown MDP}.
\end{definition}

\begin{theorem}[Probabilistic pushdown representation]
\label{thm:cfa-quant}
Under a fixed environment interpretation, the model in
Definition~\ref{def:cfa-mabs} induces the same abstract trace
distribution as $\Pr^{\Tr}_{A,\Abs}$, under the standard semantics of
probabilistic pushdown systems and pushdown MDPs.
\end{theorem}

\begin{proof}[Proof sketch]
The architecture exposes finitely many control states and a stack with
finite alphabet. Each enabled step induces either a probability
distribution over finitely many successor control/stack/action updates
or a set of such distributions under nondeterminism. This yields a
probabilistic pushdown system (or pushdown MDP) whose one-step trace
generation matches the pushforward semantics of
Definition~\ref{def:prob-trace}.
\end{proof}

\paragraph{Quantitative verification queries.}
For an unsafe region $\widehat B\subseteq\widehat S$ (or an unsafe set
of control-stack configurations), one can ask for extrema of reachability
probability, analogously to the finite-state case, but now over
pushdown models.

\section{TC Agents induce Recursively Enumerable Trace Support}
\label{sec:tca}

A \textbf{TC Agent} (Turing-complete agent) is a controller architecture
with finite control and access to an \emph{unbounded}, arbitrarily
readable and writable persistent store across steps.
Examples include an unbounded scratchpad, unrestricted file I/O used as
persistent memory, or any external store that can grow without bound
and can be accessed non-LIFO (random access) across execution.

This section states the correspondence in the same trace-centric form as
before: the set of feasible finite interaction traces is recursively
enumerable in the unbounded model, and verification in general inherits
undecidability barriers unless one bounds executions or abstracts to a
finite model.

\subsection{Architecture constraint}
\label{sec:tc-arch}

\begin{definition}[TC Agent (architecture)]
\label{def:tc-arch}
A TC Agent architecture consists of:
(i) a finite control state set $S$,
(ii) finite alphabets $\Sigma$ (perceptions) and $\Gamma$ (actions), and
(iii) an unbounded persistent store with a finite memory alphabet
$\Gamma_{\mathsf{mem}}$ and operations that allow reading and writing at
arbitrary addresses across steps (not restricted to a stack discipline).
The controller can iterate without a global, structurally enforced bound
on the number of steps.
\end{definition}

\subsection{Recursively enumerable support for interaction traces}
\label{sec:tc-support}

\begin{theorem}[Recursively enumerable trace support]
\label{thm:tc-support}
Let $A$ be a TC Agent architecture in the unbounded model
(Definition~\ref{def:tc-arch}).
Then the set of feasible finite interaction traces
$\Tr(A)\subseteq(\Sigma\times\Gamma)^*$ is recursively enumerable.
Conversely, for any recursively enumerable language
$L\subseteq(\Sigma\times\Gamma)^*$, there exists a TC Agent
architecture $A_L$ such that $\Tr(A_L)=L$.
\end{theorem}
\vspace{-1em}
\begin{proof}[Proof sketch]
For the forward direction, simulate the controller and store operations
by a Turing machine that enumerates all finite execution prefixes and
prints the corresponding interaction traces when they occur; this is a
semi-decision procedure for membership in $\Tr(A)$, hence $\Tr(A)$ is
recursively enumerable.
For the converse direction, given an r.e.\ language
$L\subseteq(\Sigma\times\Gamma)^*$ recognized by a Turing machine $M$,
construct a TC Agent that simulates $M$ on its unbounded store and emits
exactly the symbols of a trace when $M$ reaches the corresponding
configuration; feasibility of traces coincides with acceptance by $M$.
\end{proof}

\subsection{Quantitative modeling and undecidability}
\label{sec:tc-quant}

The probabilistic trace semantics (Definition~\ref{def:prob-trace})
extends to TC Agents by placing a probability measure on runs via the
policy and environment.
However, in the unbounded model, many verification questions inherit the
standard undecidability barriers of universal computation.
In practice, analysis proceeds by (i) bounding executions (step limits,
timeouts, token budgets, or storage quotas), or (ii) constructing a
finite abstraction $\Abs$ and analyzing the induced model $M_{\Abs}$
(Definition~\ref{def:abstract-model}).

When explicit bounds are imposed, the induced execution space becomes
finite-state (or at least finitely representable), yielding a model that
can be checked with bounded reachability or finite-horizon variants of
probabilistic model checking.

\section{Extension to Multi-Agent Systems}
\label{sec:mas}

The correspondence extends to multi-agent systems under standard
composition semantics.
Let $\mathcal{A}_1,\dots,\mathcal{A}_n$ be controllers with local control
states $S_1,\dots,S_n$.
A common execution semantics is \emph{interleaving}: at each global step
a scheduler selects one agent to act, that agent consumes a perception
symbol and emits an action, and the global state updates accordingly.


\textbf{Products of bounded-memory agents remain bounded-memory}
\label{sec:mas-finite}
If every $\mathcal{A}_i$ is Regular (i.e., has finite control and persistently stored memory bounded by a constant), then their composition yields a system with a finite global control-state space:
$
S_{\mathsf{global}} = S_1\times\cdots\times S_n,
$
possibly extended with finite scheduler bookkeeping.
Hence, under interleaving semantics, the composed system still belongs to the Regular class, although the state space may grow multiplicatively.
An analogous result holds for Context-Free agents under standard pushdown product constructions, provided each agent uses its own stack and the scheduler is finite-state.

\textbf{Shared unbounded memory yields TC behavior}
\label{sec:mas-shared}
The computational class changes when agents share an unbounded,
arbitrarily readable/writable persistent store.
Even if each individual agent has finite control, the composition of
finite control with a shared unbounded store can simulate a Turing
machine; The global finite control implements the machine's finite
control, and the shared store implements the tape (or random-access
memory). Under a serialized scheduler (or atomic read/write operations),
this simulation is straightforward.
Therefore, a multi-agent system of Regular controllers with shared
unbounded read/write memory is TC in the unbounded model.

\textbf{Probabilistic traces under interleaving}
\label{sec:mas-prob}
Under stochastic policies and uncertain tools, the multi-agent closed
loop induces a probability distribution over global interleaving traces.
These traces can be defined as sequences in
$(\Sigma\times\Gamma\times\{1,\dots,n\})^*$ that also record the
acting agent at each step.
All trace-based constructions above lift by including the scheduler
choices either as part of nondeterminism (adversarial scheduling) or as
part of probabilistic evolution (randomized scheduling), depending on modeling intent.

\section{Right-Sizing Agents}
\label{sec:rightsizing}

This section instantiates the architectural correspondences of
Sections~\ref{sec:reg-prob} through \ref{sec:tca} as a \emph{right-sizing}
guideline. We propose that given a task specification and a candidate framework, select
the weakest controller class whose \emph{enforced} memory interface and
control discipline can realize the required interaction behavior.
Right-sizing is stated at the architecture level rather than at the
level of an individual policy, since the policy may be stochastic and
is treated as selecting among architecturally enabled transitions.

\subsection{Decision procedure at the architecture level}
\label{sec:rightsizing-proc}

Right-sizing depends on (i) whether persistent information can grow
with execution, and (ii) how that information can be accessed across
steps. Concretely, for a controller architecture:
\begin{itemize}
  \item \textbf{Bounded persistent memory:} all information that can
  influence future control is constant-bounded and encoded into finite
  control (\autoref{sec:reg-prob}).
  \item \textbf{Stack-scoped persistent memory:} the only unbounded
  persistence is a LIFO stack, optionally with a call-return discipline
  (\autoref{sec:cfa-prob}).
  \item \textbf{Unbounded read/write persistent memory:} the controller
  can grow a store without bound and later access it in a non-stack
  manner (random access or general key-value retrieval), yielding the
  unbounded model (\autoref{sec:tca}).
\end{itemize}

Figure~\ref{fig:decisionchart} summarizes these checks. The key point is
that the classification is determined by \emph{enforced} interfaces. The
same software framework may realize different classes depending on how
it constrains memory, iteration, and tool access.

\begin{figure*}[h!]
\centering
\resizebox{1.75\columnwidth}{!}{%
\begin{tikzpicture}[node distance=1.55cm, auto, >=stealth', font=\large]
    \node[draw, rounded corners, text width=3.3cm, align=center] (start)
      {Task and interface specification\\(perceptions $\Sigma$, actions $\Gamma$)};
    \node[diamond, draw, right=of start, text width=4.0cm, align=center] (q0)
      {Does the controller require persistent information whose size can grow with execution?};
    \node[draw, rounded corners, below=of q0, text width=3.2cm, align=center, yshift=0.9cm] (regular)
      {Use \textbf{Regular}\\(finite-state)\\(\autoref{sec:reg-prob})};
    \node[diamond, draw, right=of q0, text width=4.0cm, align=center] (q1)
      {If unbounded persistence is needed, is access restricted to a LIFO stack (call-return or stack updates only)?};
    \node[draw, rounded corners, below=of q1, text width=3.2cm, align=center, yshift=0.9cm] (cfa)
      {Use \textbf{Context-Free}\\(pushdown)\\(\autoref{sec:cfa-prob})};
    \node[draw, rounded corners, right=of q1, text width=3.2cm, align=center] (tc)
      {Requires \textbf{TC}\\(unbounded read/write)\\(\autoref{sec:tca})};

    \path[->] (start) edge (q0);
    \path[->] (q0) edge node[pos=0.5, left] {No} (regular);
    \path[->] (q0) edge node[above] {Yes} (q1);
    \path[->] (q1) edge node[pos=0.5, left] {Yes} (cfa);
    \path[->] (q1) edge node[above] {No} (tc);
\end{tikzpicture}%
}
\caption{Architecture-level decision procedure for right-sizing. The checks
are phrased in terms of enforced persistent memory interfaces and access
discipline, rather than policy behavior.}
\label{fig:decisionchart}
\end{figure*}

\subsection{Illustrative mapping of patterns under explicit constraints}
\label{sec:rightsizing-mapping}

Table~\ref{tab:mapping} gives an illustrative mapping from common agent
patterns to classes, but only under explicit architectural constraints.
The purpose of the table is to emphasize that ``framework'' alone is not
the classifier; the classifier is the \emph{configuration} of memory and
control enforced by the framework.

\begin{table}[h!]
\centering
\footnotesize
\begin{threeparttable}
\caption{Illustrative mapping from patterns to classes under explicit constraints}
\vspace{-4pt}
\label{tab:mapping}
\begin{tabularx}{\linewidth}{Y Y Y}
\toprule
\textbf{Pattern / Framework} & \textbf{Constraint knob (enforced)} & \textbf{Class} \\
\midrule
Rule-based workflows (IFTTT\tnote{a}, n8n\tnote{b})
& Finite control graph; no persistent store beyond constant parameters.
& Regular \\
\midrule
Hierarchical task delegation (manager--worker)
& Stack-scoped task frames; returns restore caller context; no general
read/write store.
& Context-Free \\
\midrule
Graph-based controllers (e.g., LangGraph\tnote{c})
& \emph{If} persistent store is bounded and cycles are bounded (or
absent), controller is finite-state; \emph{if} stack-scoped frames only,
controller is pushdown.
& Regular or Context-Free \\
\midrule
ReAct-style loops~\cite{yao_react_2023}
& \emph{If} scratchpad and history are bounded and not reread beyond a
fixed window, finite-state abstraction may apply; \emph{if} the agent can
accumulate and arbitrarily reread an unbounded store, TC.
& Regular (bounded) or TC (unbounded) \\
\midrule
Auto-GPT-style agents~\cite{yang_auto-gpt_2023}
& Unbounded file or database artifacts with arbitrary reread and
iteration without structural bound.
& TC \\
\bottomrule
\end{tabularx}

\begin{tablenotes}[flushleft]
\scriptsize
\setlength{\parskip}{0pt}
\item [a] \url{https://ifttt.com/} \item [b] \url{https://n8n.io/} \item [c] \url{https://www.langchain.com/langgraph}
\end{tablenotes}
\end{threeparttable}
\end{table}

\section{Case Study: A Booking Controller}
\label{sec:casestudy}

This section illustrates right-sizing by instantiating a flight booking
task under three controller architectures that share the same external
interface but differ in their persistent memory constraints.

\textbf{Task.}
The user requests: ``book a flight to London for under \$500.''
Perceptions are tokenized symbols in $\Sigma$ (user messages and tool
outputs) and actions are tool invocations in $\Gamma$, such as
\texttt{search\_flights}, \texttt{check\_calendar}, and \texttt{book}.
Each execution induces an interaction trace in $(\Sigma\times\Gamma)^*$
(Definition~\ref{def:runs-traces}).
All three controllers invoke the same tools, but the architecture
restricts what information can persist across steps and how it can be
accessed. For analysis, each design yields an induced abstract model
$M_{\Abs}$ (Definition~\ref{def:abstract-model}) under a chosen
environment interpretation (probabilistic vs nondeterministic tool
outcomes).

\vspace{-1em}

\subsection{Regular design}
\label{sec:case-regular}

\textbf{Architecture.}
The controller has a finite set of states encoding a bounded set of
slots, for example:
\texttt{Get\_Destination}, \texttt{Get\_Date}, \texttt{Get\_Budget},
\texttt{Confirm}, \texttt{Book}.
Each slot ranges over a finite domain (or is discretized), so all
persistent information is encoded into finite control.
\textbf{Behavioral envelope.}
The feasible trace support is regular
(\autoref{thm:reg-support}).
This controller can realize traces that alternate between requesting
missing slot values and invoking tools once all slots are filled.
It does not realize traces that require unbounded accumulation of
information or hierarchical subtask nesting with arbitrary depth.
\textbf{Verification implication.}
Under a finite abstraction $\Abs$, the induced model $M_{\Abs}$ is a
finite MC/MDP (\autoref{thm:reg-quant}), which allows for probabilistic
reachability checks (e.g., bounding the probability of calling
\texttt{book} without user confirmation) rather than implying that the system is safe by construction.

\vspace{-1em}

\subsection{Context-Free design}
\label{sec:case-cfa}
\textbf{Architecture.}
The controller maintains a stack of task frames, with a designated
top-level frame \texttt{Plan\_Trip}. When the user requests ``check my
calendar before booking,'' the controller pushes a subtask frame
\texttt{Check\_Calendar}, executes it, and returns to its caller. It can
then push \texttt{Search\_Flights} and later \texttt{Book\_Flight}.
The only unbounded persistence is the LIFO stack; there is no general
read/write store.
\textbf{Behavioral envelope.}
The feasible trace support is context-free
(\autoref{thm:cfa-support}).
The stack discipline enables hierarchical plans with unbounded nesting
depth across executions, while constraining how information persists and
is revisited.
\textbf{Verification implication.}
The induced model is a probabilistic pushdown system or pushdown MDP
(\autoref{thm:cfa-quant}), which supports quantitative reachability queries
over call-return behavior.
Termination is not implied by pushdown structure; it is a property that
can be analyzed under additional assumptions (for example, bounded
retry policies, ranking arguments, or explicit step limits) or as a
bounded-execution query on a finite abstraction.
\vspace{-1em}
\subsection{TC design}
\label{sec:case-tc}

\textbf{Architecture.}
The controller can create and accumulate persistent artifacts (files,
notes, database entries) without bound and can reread them arbitrarily
across steps. It can iterate without a structurally enforced bound on
the number of tool invocations (e.g., repeated \texttt{google\_search}
followed by note-taking and refinement).
\textbf{Behavioral envelope.}
In the unbounded model, the set of feasible interaction traces is
recursively enumerable (\autoref{thm:tc-support}).
This architecture can realize traces that correspond to open-ended
research and refinement loops whose length and stored information grow
without bound.
\textbf{Verification implication.}
Quantitative analysis can still be performed on bounded executions or on
finite abstractions $\Abs$, but general unbounded verification inherits
the standard undecidability barriers of universal computation
(\autoref{sec:tca}).

\vspace{-1em}

\subsection{Summary of the right-sizing trade-off}
\label{sec:case-summary}

All three controllers share the same external tool interface, but their
memory interfaces determine which interaction traces are feasible and
which verification techniques apply.
For booking-style tasks that do not require open-ended information
accumulation, Regular or Context-Free designs can realize the relevant
trace set while yielding finite or pushdown probabilistic models
amenable to quantitative model checking.
TC designs are appropriate only when the specification explicitly
requires unbounded read/write persistence or unbounded iteration.
\vspace{-1em}

\section{Discussion}
\label{sec:discussion}

\subsection{Why Memory-Based Constraints}
Several taxonomies classify autonomy at the behavioral level, including
graded autonomy levels in automotive settings (e.g., levels of driving
automation)~\cite{SAEJ3016_2021}.
Behavioral taxonomies are descriptive and do not, by themselves, yield a
checkable transition model for verification.
In contrast, the memory interface exposed by a controller architecture
is a structural constraint in that it specifies what information can persist
across steps and how that information can be accessed.
Once these constraints are explicit, the controller can be placed in a
machine class whose induced trace semantics admits established analysis
techniques, including probabilistic model checking on a finite abstract
model $M_{\Abs}$ (Definitions~\ref{def:alpha} through \ref{def:abstract-model}).
The result is not that safety holds by construction, but that the
controller class determines which verification questions can be posed
algorithmically and which questions inherit undecidability.

\subsection{Implications for Engineering Practice}
The classification supports engineering decisions that depend on the
interaction envelope and on the available assurance techniques.

When the controller admits a finite or pushdown abstraction, safety and
reachability claims can be reduced to explicit queries on $M_{\Abs}$,
including quantitative reachability under probabilistic and
nondeterministic tool outcomes.
This supports structured assurance arguments that reference an explicit
control model, an abstraction map $\Abs$, and an interpretation of
nondeterminism.

Additionally, bounding the controller’s persistent memory interface restricts the set
of feasible traces and can reduce the space of possible behaviors.
In deployed systems, additional execution bounds (step limits, token
budgets, retry limits) can be represented directly in $M_{\Abs}$ and
analyzed as reachability or expected-cost queries.
These bounds also make it possible to compare architectures using
quantitative measures that align with operational constraints.

Finally, even when the policy component is not interpretable, architecture-level
constraints can be enforced at the tool and memory interfaces (for
example, disallowing external writes, restricting reads to a stack
frame, or mediating tool calls through an explicit controller).
This yields a separation between policy selection and allowable actions
that is compatible with the trace-based semantics.

\subsection{Implications of Correspondence}
\subsubsection{Decidability boundaries for verification queries}
The correspondence results connect controller architectures to known
machine classes and therefore to known algorithmic boundaries.

For Regular Agents, properties such as reachability and safety on the
induced finite model $M_{\Abs}$ are decidable, and quantitative variants
are addressed by probabilistic model checking of Markov chains and MDPs
over a finite state space. As for Context-Free Agents with a call-return discipline, qualitative and quantitative reachability questions can be phrased on a pushdown model
(or pushdown MDP) induced by the abstraction.
Other properties, such as language equivalence, are undecidable for
general PDAs and remain challenging even under determinism. 
Finally, For TC Agents in the unbounded model, reachability questions
inherit undecidability via standard reductions from universal
computation.
In this setting, analysis proceeds through bounded executions, finite
abstractions $\Abs$, or additional restrictions that recover decidable
subclasses.

\subsubsection{Probabilistic trace abstractions and refinement}
The semantics assigns a distribution over traces
(Definition~\ref{def:prob-trace}), and verification targets an induced
abstract probabilistic model $M_{\Abs}$
(Definitions~\ref{def:alpha}--\ref{def:abstract-model}).
Coarse abstractions can admit abstract traces that do not correspond to
any concrete execution, or can introduce spurious high-probability paths
under an MDP interpretation.
Refinement can be driven by counterexamples returned by probabilistic
model checking, by separating abstract states that conflate
control-relevant distinctions.
This yields a trace-oriented refinement loop that is compatible with
CEGAR-style workflows for nondeterministic and probabilistic systems.

\subsubsection{Planning structure and controller class}
Hierarchical planning can be represented in multiple ways depending on
what the architecture enforces.
If the architecture enforces call-return structure with stack-scoped
persistence, then a pushdown model is appropriate.
If plan depth and stored information are bounded by design, the behavior
can be encoded into finite control.
If the architecture supports unbounded accumulation and arbitrary reread
of persistent artifacts, then the unbounded model applies.

\subsection{Limitations and Threats to Validity}
This framework relies on assumptions that may not hold in all deployed
systems.
\textbf{LLM context as implicit memory.}
The policy component relies on the prompt context, 
which serves as a bounded memory whose capacity is determined 
by the context window and by how prompts are constructed in the architecture. 
The finite-state and pushdown mappings require that all control-relevant 
information provided to the policy respect this intended memory interface, 
and that information outside this interface cannot affect control decisions.

\textbf{Abstraction fidelity.}
Claims about $M_{\Abs}$ depend on whether $\Abs$ preserves the
control-relevant distinctions needed for the property of interest.
If the abstraction omits variables that influence enabled actions or
tool arguments, the induced model may be unsound for verification. 
Furthermore, while practical tools accept infinite domains (e.g., strings), 
we assume arguments are discretized via abstraction or bounded by model 
generation limits to maintain a finite alphabet $\Gamma$. 
Additionally, this framework classifies the controller’s 
persistent memory constraints, not the computational power of external tools.
Finally, Even when verification is decidable, the induced model can be large.
Practical analysis requires abstraction, compositional reasoning, and
symbolic techniques.

\textbf{Environment and tool modeling.}
Tools and environments may be nondeterministic, stochastic, or
non-stationary.
Probability estimates may change over time, and modeling uncertainty as
nondeterminism can be conservative.
The interpretation of nondeterminism (worst case, best case, or
strategy-restricted) affects reported quantitative guarantees.

\textbf{Multi-agent semantics.}
For multi-agent settings, the relationship between interleavings,
shared memory, and atomicity assumptions affects the induced trace model
and may require a more detailed operational semantics than considered in
this paper.

\section{Conclusion and Outlook}
\label{sec:conclusion}

This paper presented a memory-interface-based framework that relates
controller architectures for agentic systems to standard machine models
via probabilistic trace semantics and finite abstraction.
Under checkable constraints on persistent memory access and control
discipline, the support of the induced trace language aligns with
regular, context-free, and recursively enumerable classes for bounded
memory, stack-scoped memory, and unbounded read/write persistence,
respectively.
The framework motivates a right-sizing guideline in which the weakest
controller class is selected that can realize the required interaction traces, and
analyze the induced abstract model $M_{\Abs}$ using probabilistic model
checking when a finite abstraction is available.

Two directions follow from this formulation.
First, \emph{minimal-class synthesis}: given a task specification and an
interface model, synthesize a controller that enforces the weakest
sufficient memory interface and emit conformance evidence that the
implementation respects the interface.
Second, \emph{hybrid architectures with runtime enforcement}: mediate
tool access through finite-state or pushdown cores while permitting
higher-power components behind monitored interfaces, and analyze the
resulting system under bounded executions and abstraction.

A shared suite of reference tasks and conformance tests for memory
interfaces would enable empirical evaluation of right-sizing decisions
and support comparisons across frameworks under explicit constraints.

\newpage
\nocite{*}
\bibliographystyle{ACM-Reference-Format}
\bibliography{main}

\end{document}